\documentclass[preprint,authoryear,11pt]{elsarticle}

\usepackage{graphicx}
\usepackage{amsmath,amsfonts}
\usepackage{epstopdf}

\usepackage{url}

\usepackage{color}

\newcommand{\red}[1]{\textcolor{red}{#1}}

\usepackage{multirow}
\usepackage[caption=false,font=footnotesize]{subfig}

\usepackage{amssymb}

\newcommand{\bq}{\begin{eqnarray*}}
\newcommand{\eq}{\end{eqnarray*}}
\newcommand{\bqn}{\begin{eqnarray}}
\newcommand{\eqn}{\end{eqnarray}}

\journal{Medical Image Analysis}
\begin{document}

\begin{frontmatter}

\title{Unified Heat Kernel Regression for Diffusion, Kernel Smoothing and Wavelets on Manifolds and Its Application to Mandible  Growth Modeling in CT Images}

\author[vocal]{Moo K. Chung} 
\author[NUS]{Anqi Qiu}
\author[bcs]{Seongho Seo}
\author[vocal]{Houri K. Vorperian}
\address[vocal]{
University of Wisconsin, Madison} 
\address[NUS]{National University of Singapore}
\address[bcs]{
Seoul National University, Korea\\
\vspace{0.5cm}
\red{\tt mkchung@wisc.edu}}

\begin{abstract}
We present a novel kernel regression framework for smoothing scalar surface data using the Laplace-Beltrami eigenfunctions. Starting with the heat kernel constructed from the eigenfunctions, we formulate a new bivariate kernel regression framework as a weighted eigenfunction expansion with the heat kernel as the weights. The new kernel method is mathematically equivalent to isotropic heat diffusion, kernel smoothing and recently popular diffusion wavelets.
The numerical implementation is validated on a unit sphere using spherical harmonics. As an illustration, the method is applied to characterize the localized growth pattern of mandible surfaces obtained in CT images between ages 0 and 20 by regressing the length of displacement vectors with respect to a surface template.
\end{abstract}


\end{frontmatter}

\section{Introduction}
\label{sec:introduction}

In medical imaging, anatomical surfaces extracted from MRI and CT are often represented as triangular meshes. The image segmentation and surface extraction processes themselves are likely to introduce noise to the mesh coordinates. It is therefore 
imperative to reduce the mesh noise while preserving the geometric details of the anatomical structures for various applications. 

Diffusion equations have been widely used in image processing as a form of noise reduction 
since 1990 \citep{perona.1990}. Numerous techniques have been developed for surface fairing and mesh regularization \citep{sochen.1998,malladi.2002,tang.1999,taubin.2000} and  surface data smoothing  \citep{andrade.2001,chung.2001.diffusion,chung.2004.isbi,cachia.tmi.2003, cachia.mia.2003, chung.2005.IPMI, joshi.2009}. Isotropic heat diffusion on surfaces has been introduced in brain imaging for subsequent statistical analysis involving the random field theory (RFT) that assumes an isotropic covariance function as a noise model \citep{andrade.2001,chung.2004.isbi,cachia.tmi.2003,cachia.mia.2003}. Since then, isotropic diffusion has been the standard smoothing technique. 

Iterated kernel smoothing has been  another widely used method in approximately solving diffusion equations on surfaces \citep{chung.2005.IPMI,han.2006}. It is often used  in smoothing anatomical surface data including cortical curvatures \citep{luders.ni.2006,gaser.2006}, cortical thickness maps \citep{luders.hbm.2006,bernal.2008}, hippocampus surfaces \citep{shen.CSB.2006,zhu.2007} and magnetoencephalography (MEG)  \citep{han.2007} and functional-MRI \citep{hagler.2006,jo.2007} data on the brain surface. Due to its simplicity, it is the most widely used form of surface data smoothing in brain imaging. In iterated kernel smoothing, kernel weights are spatially adapted to follow the shape of the heat kernel in a discrete fashion along a manifold. In the tangent space of the manifold, the heat kernel with a small bandwidth can be approximated linearly using the Gaussian kernel. The heat kernel with a large bandwidth is then constructed by iteratively applying the Gaussian kernel with the small bandwidth. However, this process compounds the linearization error at each iteration.

We propose a new kernel regression framework that constructs the heat kernel analytically using the eigenfunctions of the Laplace-Beltrami (LB) operator, avoiding the need for the linear approximation used by \citet{chung.2005.IPMI} and \citet{han.2006}. Although a few studies have introduced the heat kernel in computer vision and machine learning, they mainly used the heat kernels to compute shape descriptors  or to define a multi-scale metric \citep{belkin.2006,sun.2009,bronstein.2010, deGoes.2008}. These studies did not use the heat kernels in regressing functional data on manifolds. This is the first study 
to use the heat kernel in the form of regression for the subsequent statistical analysis. There have been significant developments in kernel methods in the machine learning community \citep{bernhard.2002,nilsson.2007,shawe.2004,steinke.2008,yger.2011}. However, to the best of our knowledge, the heat kernel has never been used in such frameworks. Most kernel methods in machine learning deal with the linear combination of kernels as a solution to penalized regressions. On the other hand, our kernel regression framework does not have a penalized cost function. 

Wavelets have recently been popularized for surface and graph data. For instance, spherical wavelets were used on brain surface data already mapped onto a sphere  \citep{nain.2007,bernal.2008}. Since the wavelet basis has local supports in both space and scale, the wavelet coefficients provide shape features that describe local shape variation at a variety of scales and spatial locations. However, spherical wavelets require a smooth mapping from the surface to a unit sphere, thus introducing a serious metric distortion that compounds subsequent statistical parametric maps (SPM). Furthermore, such basis functions are only orthonormal for data defined on the sphere and result in a less parsimonious  representation for data defined on other surfaces compared to the intrinsic LB-eigenfunction expansion \citep{seo.2011.ISBI}. To remedy the limitations of spherical wavelets, the diffusion wavelet transform on graph data structures has been proposed \citep{antoine.2010,coifman.2006,hammond.2011,kim.2012.NIPS}. 

The primary methodological contribution of this study is the establishment of a unified regression framework that combines the diffusion-, kernel- and wavelet-based methods in a coherent mathematical fashion for scalar data defined on manifolds. We unify the apparent differences between the methods while providing detailed theoretical justifications. This paper extends the work by \citet{kim.2011.PSIVT}, which  introduced heat kernel smoothing to smooth out surface noise in the hippocampus and amygdala. Although the idea of diffusion wavelet transform for surface meshes was explored by  \citet{kim.2012.NIPS}, the relationship between the wavelet transform and the proposed kernel regression was not investigated. For the first time, the mathematical equivalence between the two constructs is explained. 

The proposed kernel regression framework was subsequently applied in characterizing the growth pattern of the mandible surfaces obtained in CT and identifying the regions of the mandible that show the most significant localized growth. The length of the displacement vector field was regressed over the mandible surface to increase the signal-to-noise ratio (SNR) and hence statistical sensitivity. To our knowledge, this is the first growth modeling of the mandible surface in a continuous fashion without using anatomic landmarks.

\section{Methods}
\subsection{Isotropic Diffusion on Manifolds}
Consider a functional measurement $Y(p)$ observed at each point $p$ on a compact manifold $\mathcal{M} \subset \mathbb{R}^3$.
We assume the following linear model on $Y$:
 \begin{equation}
 \label{model}
Y(p) = \theta(p) +\epsilon(p),
\end{equation}
where $\theta(p)$ is the unknown mean signal to be estimated and $\epsilon(p)$ is a zero-mean Gaussian random field. We may assume further  $Y \in L^2(\mathcal{M})$, the space of square integrable functions on $\mathcal{M}$ with the inner product:
 \begin{equation} 
\langle f, g \rangle = \int_{\mathcal{M}} f(p) g(p) \; d\mu(p),
\end{equation}
where $\mu$ is the Lebesgue measure such that $\mu(\mathcal{M})$ is the total area of $\mathcal{M}$.

Imaging data such as electroencephalography (EEG), magnetoencephalography (MEG) \citep{han.2007} and functional-MRI \citep{hagler.2006,jo.2007}, and anatomical data  such as cortical curvatures \citep{luders.ni.2006,gaser.2006}, cortical thickness \citep{luders.hbm.2006,bernal.2008} and surface coordinates \citep{chung.2005.IPMI} can be considered as 
functional measurements. Functional measurements are expected to be noisy and require filtering to boost signal.

Surface measurements have often been filtered using the isotropic diffusion equation of the form \citep{andrade.2001,chung.2001.diffusion,cachia.tmi.2003,rosenberg.1997}
\bqn 
	\frac{\partial f}{\partial \sigma} = \Delta f,\;
	f(p,\sigma = 0)=Y(p)
\label{eq:isotropic.diff},
\eqn
where $\Delta$ is the Laplace-Beltrami operator defined on manifold $\mathcal{M}$.
The diffusion time $\sigma$ controls the amount of smoothing. It can be shown that
the unique solution of (\ref{eq:isotropic.diff}) is given by a kernel convolution 
as follows. A Green's function or a fundamental solution of the Cauchy problem (\ref{eq:isotropic.diff}) is given by the solution of the following equation:
\bqn 
	\frac{\partial f}{\partial \sigma} = \Delta f,\;
	f(p,\sigma=0)=\delta(p), 
\label{eq:smoothing-fund} 
\eqn
where $\delta$ is the Dirac delta function. The heat kernel $K_{\sigma}$ is a Green's function of (\ref{eq:smoothing-fund}) \citep{evans.1998}, i.e.,
$$ 	\frac{\partial K_{\sigma}}{\partial \sigma} = \Delta K_{\sigma},\;
	K_{\sigma}(p,\sigma=0)=\delta(p).$$
Since the differential operators are linear in  (\ref{eq:smoothing-fund}), we can further convolve 
the terms with the initial data $Y$ such that 
$$ \frac{\partial }{\partial \sigma} (K_{\sigma} * Y )  = \Delta (K_{\sigma}  * Y ),  \; K_{\sigma}*Y (p,\sigma=0) = Y(p),$$
where $$K_{\sigma} * Y(p) = \int_{\mathcal{M}} K_{\sigma}(p,q) Y(q) \; d\mu(q).$$
Hence $K_{\sigma}*Y$ is a solution of (\ref{eq:isotropic.diff}).

\begin{figure}[t]
\centering
\includegraphics[width=0.8\linewidth]{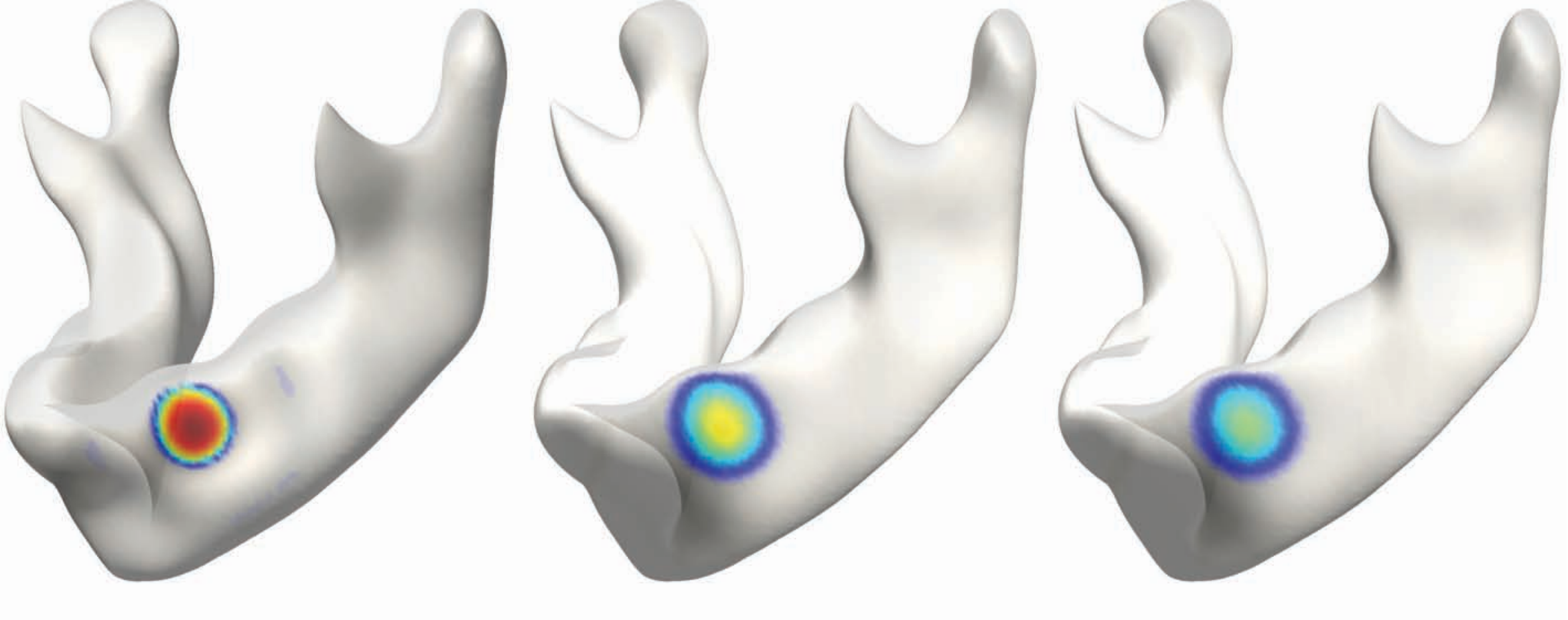} 
\caption{The heat kernel shape with bandwidths 0.025 (left), 1.25 (middle) and 5 (right) on a mandible surface.  The level sets of the heat kernel form geodesic circles.}
\label{fig:9_hk_shape}
\end{figure}

\subsection{Diffusion Smoothing}
Isotropic diffusion \eqref{eq:isotropic.diff} has been 
solved by various numerical techniques \citep{chung.2001.thesis,
andrade.2001,cachia.tmi.2003,cachia.mia.2003,chung.2004.isbi}. 
For diffusion smoothing, the diffusion equation needs to be discretized using the cotan formulation \citep{chung.2001.thesis,chung.2004.isbi,qiu.2006}. Since there are many different cotan formulations, we follow the formulation introduced in \cite{chung.2001.thesis}. The diffusion equation (\ref{eq:isotropic.diff}) is discretized as
\begin{equation}\label{eq:discreteDiff3}
	\frac{\partial \mathbf{f}}{\partial \sigma} = -\mathbf{A}^{-1} \mathbf{C} \mathbf{f},
\end{equation}
where $\mathbf{f} = (f(p_1, \sigma), \cdots, f(p_n, \sigma))'$ is the vector of measurements over all mesh vertices at time $\sigma$. $\mathbf{A}=(A_{ij})$ is the stiffness matrix and $\mathbf{C} = (C_{ij})$ is the global coefficient matrix, which is the assemblage of individual element coefficients. The sparse matrices $\mathbf{A}$ and $\mathbf{C}$ are explicitly given as follows.

Let $T_{ij}^-$ and $T_{ij}^+$ denote two triangles sharing the vertex $p_i$ and its neighboring vertex $p_j$ in a mesh. 
Let the two angles opposite to the edge containing $p_i$ and $p_j$ be $\phi_{ij}$ and $\theta_{ij}$ respectively for $T_{ij}^+$ and $T_{ij}^-$. 
The off-diagonal entries of the stiffness matrix are
$$A_{ij} = \frac{1}{12}  \big( |T_{ij}^+| + |T_{ij}^-| \big)$$
if $p_i$ and $p_j$ are adjacent and $A_{ij}=0$ otherwise. $| \cdot |$ denotes the area of a triangle. 
The diagonal entries are summed as $A_{ii} = \sum_{j=1}^n A_{ij}.$
The off-diagonal entries of the global coefficient matrix are
$$C_{ij} = -\frac{1}{2} ( \cot \theta_{ij} + \cot \phi_{ij})$$
if $p_i$ and $p_j$ are adjacent and $C_{ij}=0$ otherwise. The diagonal entries are similarly given as the sum $C_{ii} =  - \sum_{j=1}^n  C_{ij}.$

The ordinary differential equation (\ref{eq:discreteDiff3}) is then further discretized at each point using the forward finite difference scheme:

\begin{equation}\label{eq:fdm}
	{\bf f(}p_i,\sigma_{n+1}) = {\bf f}(p_i,\sigma_{n}) + (\sigma_{n+1}-\sigma_{n})\widehat{\Delta} f(p_i, \sigma_{n}) ,
\end{equation}
where $\widehat{\Delta} f(p_i,\sigma_{n})$ is the estimated Laplacian obtained from the $i$-th row of $-\mathbf{A}^{-1} \mathbf{C} \mathbf{f}$.
For the forward Euler scheme to converge,  we need to have a sufficiently small  step size 
 $\Delta \sigma = \sigma_{n+1}-\sigma_{n}$ \citep{chung.2001.thesis}.

\subsection{Iterated Kernel Smoothing}
The diffusion equation (\ref{eq:isotropic.diff}) can be approximately solved by iteratively performing Gaussian kernel smoothing \citep{chung.2005.IPMI}. The weights of the kernel are spatially adapted to follow the shape of the heat kernel along a surface mesh. Heat kernel smoothing with a large bandwidth can be broken into iterated smoothing with smaller bandwidths \citep{chung.2005.IPMI}:
\begin{equation}\label{eq:hk_scale}
K_{m\sigma} \ast Y =\underbrace{ K_{\sigma} * \cdots * K_{\sigma} }_{m \; \mbox{times}}* Y.
\end{equation}
Then using the parametrix expansion \citep{rosenberg.1997,wang.1997}, we approximate the heat kernel with the small bandwidth locally using the Gaussian kernel:
\begin{equation}
	K _{\sigma}(p, q) = \frac{1}{\sqrt{4\pi\sigma}} \exp [ -\frac{d^2(p,q)}{4\sigma} ][1+O(\sigma^2)],
\end{equation}
where $d(p,q)$ is the geodesic distance between $p$ and $q$. For sufficiently small bandwidth $\sigma$, all 
kernel weights are concentrated near the center, so the first neighbors of a given mesh vertex are sufficient for approximation. Unfortunately, this approximation compounds error at each 
 iteration. For numerical implementation, we used the normalized truncated kernel given by
\begin{equation}
	W _{\sigma}(p, q_i) = \frac{\exp \big[ -\frac{d^2(p,q_i)}{4\sigma} \big]}{\sum_{j=0}^r\exp \big[ -\frac{d^2(p,q_j)}{4\sigma} \big]},
\end{equation}
where $q_1, \cdots, q_r$ are $r$ neighboring vertices of $p=q_0$. 
Denote the truncated kernel convolution as 
\begin{equation}\label{eq:itr}
W_{\sigma} \ast Y(p) = \sum_{i=0}^{r} W_{\sigma}(p,q_i)Y(q_i).
\end{equation}
The iterated heat kernel smoothing is then defined as
$$W_{m\sigma} \ast Y(p) = \underbrace{ W_{\sigma} * \cdots * W_{\sigma} }_{m \; \mbox{times}}* Y(p).$$

\subsection{Heat Kernel Regression}
We present a new regression framework for solving the isotropic diffusion equation
 (\ref{eq:isotropic.diff}). Let $\Delta$ be the Laplace-Beltrami operator on $\mathcal{M}$. Solving the eigenvalue equation
\bqn 
\Delta \psi_j = -\lambda \psi_j \label{eq:eigensystem},
\eqn 
we order the eigenvalues 
$$0 =  \lambda_0  < \lambda_1 \leq \lambda_2 \leq \cdots $$ 
and corresponding eigenfunctions $\psi_0, \psi_1, \psi_2, \cdots$ \citep{rosenberg.1997,chung.2005.IPMI,levy.2006,shi.2009}. The first eigenvalue and eigenfunction are trivially given as $\lambda_0=0$  and $\psi_0= 1/\sqrt{\mu(\mathcal{M})}$. It is possible to have multiple eigenfunctions corresponding to the same eigenvalue. 

The eigenfunctions $\psi_j$ form an orthonormal basis in $L^2(\mathcal{M})$. There is extensive literature on the use of eigenvalues and eigenfunctions of the Laplace-Beltrami operator in medical imaging and computer vision \citep{levy.2006,qiu.2006,reuter.2009.CAD,reuter.2010.IJCV,zhang.2007,zhang.2010}. The eigenvalues have been used in caudate shape discriminators \citep{niethammer.2007}. Qiu et al. used eigenfunctions in constructing splines on cortical surfaces \citep{qiu.2006}. Reuter used the topological features of eigenfunctions \citep{reuter.2010.IJCV}. Shi et al. used the Reeb graph of the second eigenfunction in shape characterization and landmark detection in cortical and subcortical structures \citep{shi.2008.miccai,shi.2009}. Lai et al. used the critical points of the second eigenfunction as anatomical landmarks for colon surfaces \citep{lai.2010}. Since the direct application of eigenvalues and eigenfunctions as features of interest is beyond the scope of this paper, we will not pursue the issue in detail here.

\begin{figure}[t]
\centering
\includegraphics[width=1\linewidth]{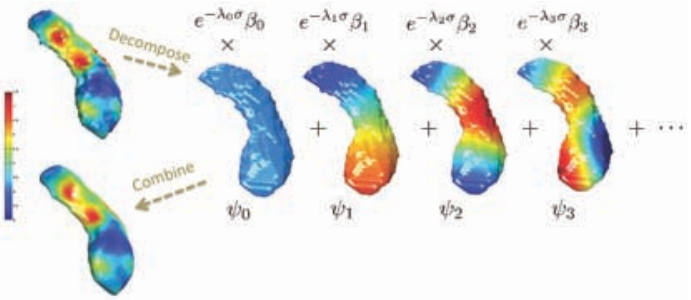}  
\caption{Schematic of heat kernel smoothing. Given functional data on a surface, we first compute the eigenfunctions $\psi_j$ and the Fourier coefficients $\beta_j$. Then, we combine all the terms and reconstruct the functional signal back.}
\label{fig:schematic-hk}
\end{figure}

Using the eigenfunctions, the {\em  heat kernel}  $K _{\sigma}(p,q)$ is  
defined as 
\begin{equation}\label{eq:hk}
	K _{\sigma}(p,q) = \sum_{j=0}^{\infty} e^{-\lambda_j \sigma} \psi_j(p) \psi_j (q),
\end{equation}
where $\sigma$ is the bandwidth of the kernel. 
Figure \ref{fig:9_hk_shape} shows examples of the heat kernel with different bandwidths. 
Then the {\em heat kernel regression} or {\em heat kernel smoothing} of functional measurement $Y$ is defined 
as
\begin{equation}\label{eq:al_smt}
K_{\sigma} \ast Y(p) = \sum_{j=0}^{\infty} e^{-\lambda_j \sigma} \beta_j \psi_j (p),
\end{equation}
where $\beta_j = \langle Y,\psi_j \rangle$ are Fourier coefficients \citep{chung.2005.IPMI} (Figure \ref{fig:schematic-hk}). The kernel smoothing $K_{\sigma}*Y$ is taken as an estimate for the unknown mean signal $\theta$ in (\ref{model}). 
The degree of truncation of the series expansion can be automatically determined using the forward model selection procedure.

Unlike previous approaches to heat diffusion \citep{andrade.2001,chung.2004.isbi,joshi.2009,tasdizen.osher.2006}, the proposed method avoids the direct numerical discretization of the diffusion equation. Instead we discretize the basis functions of the manifold $\mathcal{M}$ by solving for the eigensystem (\ref{eq:eigensystem}) to obtain $\lambda_j$ and $\psi_j$.

\subsection{Diffusion Wavelet Transform} 

We can establish the equivalence between the proposed kernel regression and recently popular diffusion wavelets. This mathematical equivalence eliminates the need for constructing wavelets using complicated computational machinery as has often been done in previous studies \citep{antoine.2010,hammond.2011,kim.2012.NIPS}, and offers a simpler but more unified alternative.  

Consider a wavelet basis $W_{\sigma,q}(p)$ obtained from a mother wavelet $W$ with scale and translation parameters $\sigma$ and $q$ respectively in a Euclidean space:
$$W_{\sigma, q}(p) = \frac{1}{\sigma}W \big(\frac{p-q}{\sigma} \big).$$
Generalizing the idea of scaling a mother wavelet in Euclidean space to a curved surface is trivial. The difficulty arises when one tries to translate a mother wavelet on a curved surface since it is unclear how to define translation along the surface. If one tries to modify  the existing spherical wavelet framework to an arbitrary surface \citep{nain.2007,bernal.2008}, one immediately encounters the problem of establishing regular grids on an arbitrary surface. 
Recent work based on diffusion wavelets bypass this problem by taking the bivariate kernel as a mother wavelet \citep{antoine.2010,hammond.2011,mahadevan.2006, kim.2012.NIPS}.

For some scale function $g$ that satisfies the admissibility conditions, 
 diffusion wavelet $W_{\sigma,q}(p)$ at position $p$ and scale $\sigma$ is given by
$$W_{\sigma,q}(p) = \sum_{j=0}^k g(\lambda_j \sigma)\psi_j(p)\psi_j(q),$$
where $\lambda_j$ and $\phi_j$ are eigenvalues and eigenfunctions of the Laplace-Beltrami operator. The wavelet transform  
is then given by 
\bqn \langle W_{\sigma,q}, Y \rangle = \int_{\mathcal{M}} W_{\sigma,q}(p) Y(p) \;d\mu(p). \label{eq:WTR} \eqn
By letting
$g(\lambda_j \sigma ) = \exp (- \lambda_j \sigma)$, 
we have the heat kernel as the wavelet, i.e.,
$$W_{\sigma,p}(q) = K_{\sigma}(p,q),$$
The bandwidth $\sigma$ of the heat kernel is the scale parameter, while the translation is achieved by shifting one argument in the bivariate heat kernel. Subsequently, wavelet transform (\ref{eq:WTR}) can then be rewritten as
\bqn \langle W_{\sigma,p}, Y \rangle = \sum_{j=0}^k e^{-\lambda_j  \sigma} \beta_j \psi_j(q) \label{eq:WT} \eqn
with $\beta_j = \langle Y, \psi_j \rangle$. The expression (\ref{eq:WT}) is 
the finite truncation of the heat kernel regression in (\ref{eq:hk}). Hence, diffusion wavelet analysis can be simply performed within the proposed heat kernel regression framework without any additional wavelet machinery. We  therefore do not distinguish between the heat kernel regression and the diffusion wavelet transform. 

\begin{figure}
  \centering
\includegraphics[width=0.7\linewidth]{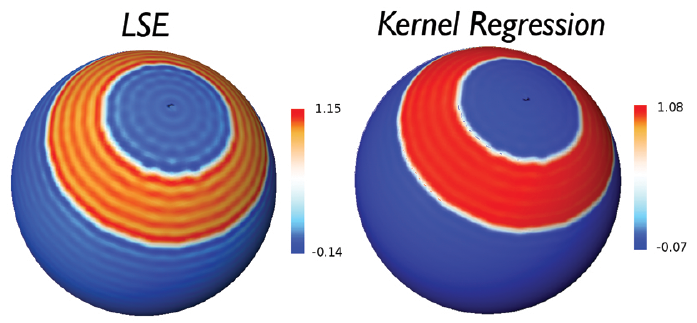}
  \caption{\footnotesize Gibbs phenomenon (ringing artifacts) is visible in the spherical harmonic series expansion with degree 78 via the least squares estimation (LSE) of the step function defined on a sphere. In contrast, the heat kernel regression with the same degree and bandwidth 0.0001 shows fewer visible artifacts.}
\label{fig:midus-gibbs}
\end{figure}

Although the heat kernel regression is constructed using the global basis functions $\psi_j$, surprisingly the kernel regression at each point $p$ coincides with the wavelet transform at that point. Hence, it also inherits  
the localization property of wavelets at that point. This is clearly demonstrated in the simulation study shown in Figure \ref{fig:midus-gibbs}, where a step function of value 1 in the circular band $1/8 < \theta < 1/4$ (angle from the north pole) and of value 0 outside of the band was constructed. Note that, on a sphere, the Laplace-Beltrami operator is the spherical Laplacian and its eigenfunctions are spherical harmonics $Y_{lm}$ of degree $l$ and order $m$. Then the step function was reconstructed using the spherical harmonic series expansion
$$Y(p) = \sum_{l=0}^{L} \sum_{m=-l}^l \beta_{lm}Y_{lm}(p),$$
where the spherical harmonic coefficients $\beta_{lm}  = \langle Y, Y_{lm} \rangle$ were obtained by the least squares estimation (LSE). 
On the unit sphere, we used the heat kernel regression of the form
$$Y(p) = \sum_{l=0}^{L} \sum_{m=-l}^l  e^{-l(l+1)\sigma}\beta_{lm}Y_{lm}(p)$$ 
with small bandwidth $\sigma=0.0001$ and degree $L=78$. The spherical harmonic expansion 
shows severe ringing artifacts compared to the kernel regression, which inherits the localization power of wavelets.  Thus the Gibbs phenomenon was not significantly visible.

\subsection{Parameter Estimation in Heat Kernel Regression}

\label{sec:FEM}
Since the closed form expression for the eigenfunctions of the Laplace-Beltrami operator on an arbitrary surface is unknown, the eigenfunctions are numerically computed by discretizing the Laplace-Beltrami operator. To solve the eigensystem (\ref{eq:eigensystem}), we need to discretize it on mandible triangular meshes using the cotan formulation \citep{chung.2001.thesis,chung.2004.isbi,shi.2009,qiu.2006,levy.2006,reuter.2006.CAD,reuter.2009.CAD,rustamov.2007,zhang.2007,vallet.2008,wardetzky.2008}.

\begin{figure}[t]
\centering
\includegraphics[width=1\linewidth]{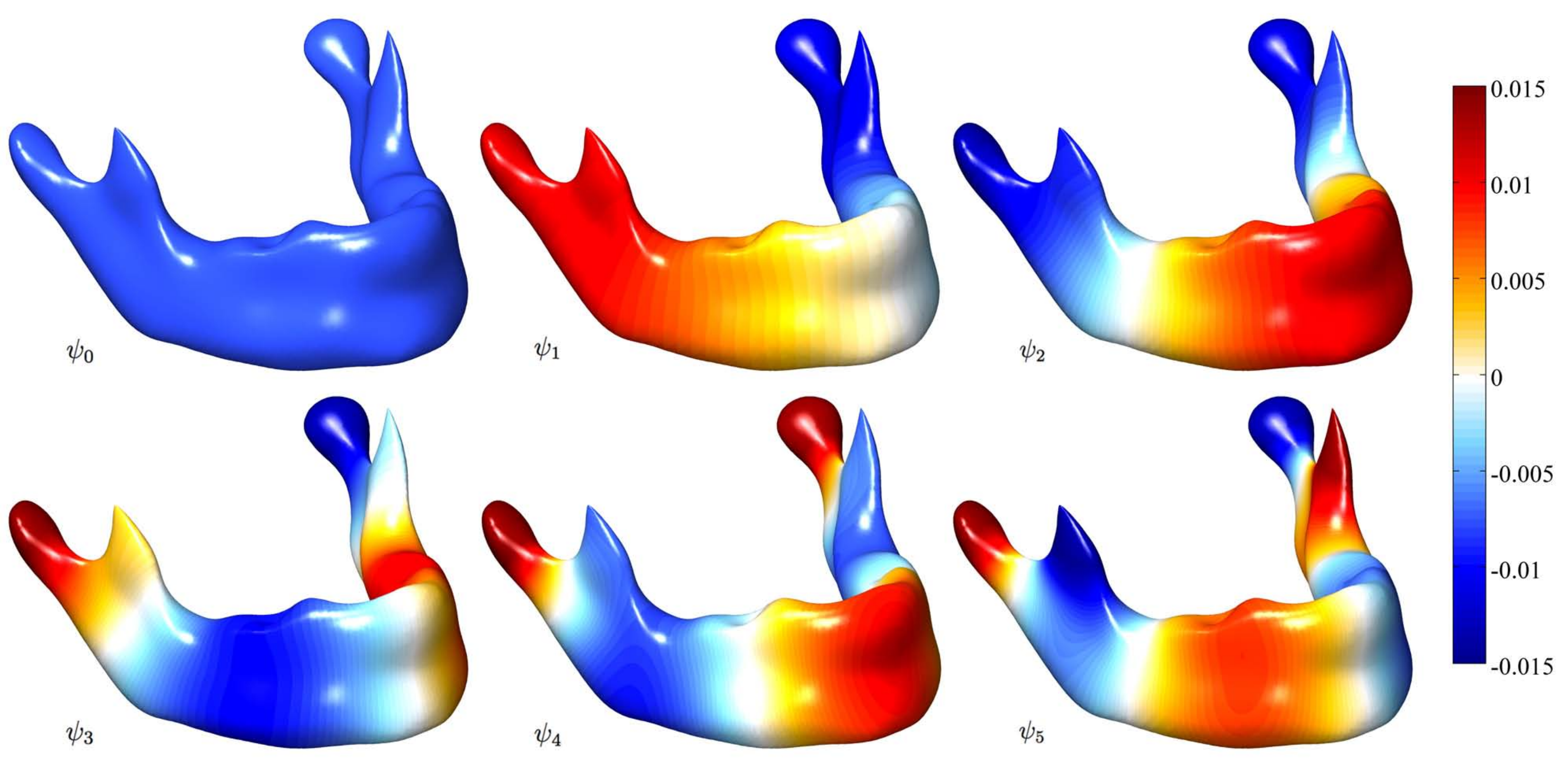}  
\caption{Eigenfunctions of various degrees for a sample mandible surface. The eigenfunctions are projected on the surface smoothed by the proposed heat kernel smoothing with bandwidth $\sigma=0.5$ and degree $k=132$. The smoothed surface is obtained by heat kernel smoothing applied to the coordinates of the surface mesh with the same parameter while preserving the topology of mesh.
The first eigenfunction is simply $\psi_0 = 1/\sqrt{\mu(\mathcal{M})}$. The color scale is thresholded for better visualization. }
\label{fig:2_md_eigfs}
\end{figure}

 Among the many different cotan formulations used in computer vision and medical image analysis, we used the formulation given in \citet{chung.2001.thesis} and \citet{qiu.2006}. It requires discretizing (\ref{eq:eigensystem}) as the following generalized eigenvalue problem:
\begin{equation}\label{eq:gep1}
	\mathbf{C\psi} = \lambda \mathbf{A\psi},
\end{equation}
where the global coefficient matrix $\mathbf{C}$ is the assemblage of individual element coefficients 
and $\mathcal{A}$ is the stiffness matrix. We solved (\ref{eq:gep1}) using the {\em Implicitly Restarted Arnoldi Method} \citep{hernande.2006,lehoucq.1998} without consuming large amounts of memory and time for sparse entries. Figure~\ref{fig:2_md_eigfs} shows the first few eigenfunctions for a mandible surface.

Once we obtain the eigenfunctions numerically, we estimate the kernel regression parameters $\beta_j$ using the least squares estimation (LSE) technique. Note $\beta_j = \langle Y, \psi_j \rangle$, the Fourier coefficients with respect to basis $\psi_j$. 
$\beta_0, \beta_1, \cdots, \beta_k$ are then estimated simultaneously by minimizing the sum of squared residual: 
 \begin{equation} \label{eq:lse}
       \arg\min_{\beta_0, \cdots, \beta_k} \Big\Vert Y(p)-  \sum_{j=0}^{k}  \beta_j \psi_j (p) \Big\Vert^2.
\end{equation}
The least squares method  is often used in estimating the coefficients in spherical harmonic expansion \citep{shen.2004,styner.2006,chung.2008.sinica}.
Suppose we have $n$ mesh vertices $p_1, \cdots, p_n$. Let  
$${\bf Y} = (Y(p_1), \cdots, Y(p_n))'$$ be the surface measurements over all $n$ vertices. Denote the $j$-th eigenfunction evaluated at $n$ vertices as
$$\mathbf{\Psi}_j= (\psi_j(p_1), \cdots, \psi_j(p_n))'.$$
The minimum in (\ref{eq:lse}) is achieved when
\bqn 
\mathbf{Y} = \mathbf{\Psi}\boldsymbol{\beta},\label{eq:normal}
\eqn
where $\mathbf{\Psi} = (\mathbf{\Psi}_0, \cdots, \mathbf{\Psi}_{k})$ is the matrix of size $n \times (k+1)$. 
The LSE estimation of coefficients $\boldsymbol{\beta}$ is then given by
 \begin{equation} \label{eq:ff}
	\boldsymbol{\widehat{\beta}} =  (\mathbf{\Psi}^\prime \mathbf{\Psi})^{-1}\mathbf{\Psi}^\prime \mathbf{Y}.
\end{equation}
Since it is expected that the number of mesh vertices is substantially larger than the number of eigenfunctions to be used, $\mathbf{\Psi}^\prime \mathbf{\Psi}$ is well conditioned and invertible. The numerical implementation is available at \url{http://www.stat.wisc.edu/~mchung/mandible} with the full data set used in the study.

\subsection{Random Field Theory (RFT)}

Once we have smoothed functional data on a surface, we apply the statistical parametric mapping (SPM) framework for analyzing and visualizing statistical tests performed on the template surface that is often used in structural neuroimaging studies \citep{andrade.2001,lerch.2005.ni,wang.2010,worsley.1995,yushkevich.2008}. Since test statistics are constructed over all mesh vertices on the mandible, multiple comparisons need to be accounted for using the RFT \citep{taylor.2007,worsley.1995,worsley.2004}. The RFT assumes the measurements to be a smooth Gaussian random field.  Heat kernel smoothing will make data smoother and more Gaussian and enhance
 the SNR \citep{chung.2005.IPMI}.  The proposed kernel regression framework can then be naturally integrated into the RFT-based statistical inference approach \citep{taylor.2007,worsley.2004,worsley.1995}. 
 
We assume $\theta$ is an unknown group level signal and $\epsilon$ is a zero-mean unit-variance Gaussian random field in (\ref{model}).  The model assumptions are not as restrictive as it seems since we can always normalize the data in this fashion. We further assume the random field $\epsilon$ is the convolution of heat kernel $K_{s}$ on Gaussian white noise $W$ with bandwidth $s$, i.e., $\epsilon (p) = K_{s}*W(p)$. Previously, the smoothness of noise, i.e., kernel bandwidth $s$, was estimated using the RESEL (resolution element) technique, which requires estimating the quantity ${\bf Var} \left[\partial \epsilon(p)/ \partial p \right]$ along mesh surfaces, which can introduce a bias \citep{worsley.1999,hayasaka.2004,kilner.2010}. Thus, surface data is often smoothed with bandwidth $\sigma$ that is sufficiently larger than $s$ so that any high frequency noise smaller than $\sigma$ is masked out. This provides a motivation for developing the heat kernel regression framework.

In (\ref{model}), we are interested in determining the significance of $\theta$, i.e.,
\bqn H_0: \theta(p) &=& 0 \mbox{ for all } p \in \mathcal{M} \nonumber \\
\mbox{ vs. } \;
H_1: \theta(p) &>& 0 \mbox { for some } x \in \mathcal{M}. \label{eq:hypothesis} \eqn
Note that any point $p_0$ that gives $\theta(p_0) > 0$ is considered as signal. The hypothesis
(\ref{eq:hypothesis}) is an infinite dimensional multiple comparisons problem for continuously indexed hypotheses over the manifold $\mathcal{M}$. The underlying group level signal $h$ is estimated using the proposed heat kernel regression. Subsequently, a test statistic is often given by a T- or F-field $Y(p)$  \citep{worsley.2004,worsley.1995}.

The multiple comparisons corrected $p$-value is then computed through by the RFT \citep{adler.1981,cao.2001,taylor.2007,worsley.2003}. For the $F$-field $Y$ with $\alpha$ and $\beta$ degrees of freedom defined on 2D manifolds $\mathcal{M}_F$, it is known that
\bqn P \Big(\sup_{p \in \mathcal{M}_F} Y(p) > h \Big) \approx  
\mu_2(\mathcal{M}_T)\rho_2(h) + \mu_0(\mathcal{M}_F)\rho_0(h)  \label{eq:deform-ECexpansion}\eqn 
for a sufficiently large threshold $h$, where $\mu_d(\mathcal{M}_F)$ is the $d$-th
Minkowski functional of $\mathcal{M}_F$ and
$\rho_d$ is the $d$-th Euler characteristic (EC) density of
$Y$ \citep{worsley.1998}. 
The Minkowski functionals are given by 
\bq \mu_2(\mathcal{M}_T) &=& \mbox{area}( \mathcal{M}_T)/2, \\
\mu_0(\mathcal{M}_T) &=&  \chi(\mathcal{M}_T) =2.
\eq 
The EC-density
for $F$-field is then given by 
\bq 
\rho_2&=&\frac{1}{4\pi\sigma^2} {{\Gamma ({{\alpha +\beta-2}\over 2})}\over
{\Gamma ({\alpha \over 2})\Gamma ({\beta \over 2})}}
\left({{\alpha h}\over \beta }\right)^{(\alpha-2) \over 2}
\left(1+{{\alpha h}\over \beta }\right)^{-{(\alpha +\beta-2)\over 2}} \left[(\beta -1){{\alpha h}\over \beta}-(\alpha-1)\right]\\
\rho_0 &=& 1 - P(F_{\alpha, \beta} \leq h),
\eq
where $P(F_{\alpha, \beta} \leq h)$ is the cumulative distribution function of $F$-stat with $\alpha$ and $\beta$ degrees of freedom. The second order term $\mu_2(\mathcal{M}_T)\rho_2(h)$ dominates the expression (\ref{eq:deform-ECexpansion}) and it explicitly has the bandwidth $\sigma$ of the kernel regression, thus incorporating the proposed kernel framework into the RFT.

\section{Experiments}
\subsection{CT Image Preprocessing}
We applied the proposed smoothing method to CT images of mandibles obtained from several different models of GE multi-slice helical CT scanners. The CT scans were acquired directly in the axial plane with $1.25$ mm slice thickness, matrix size of $512 \times 512$ and $15$--$25$ cm 
field of view (FOV). Image resolution varied as voxel size ranged from 0.25 mm$^3$ to 0.49 mm$^3$ as determined by the ratio of FOV divided by the matrix. CT scans were converted to DICOM format and Analyze $8.1$ software package (AnalyzeDirect, Inc., Overland Park, KS) was then used in segmenting binary mandible structure based on histogram thresholding.

\begin{figure}[t]
\begin{center}
\includegraphics[width=0.75\linewidth]{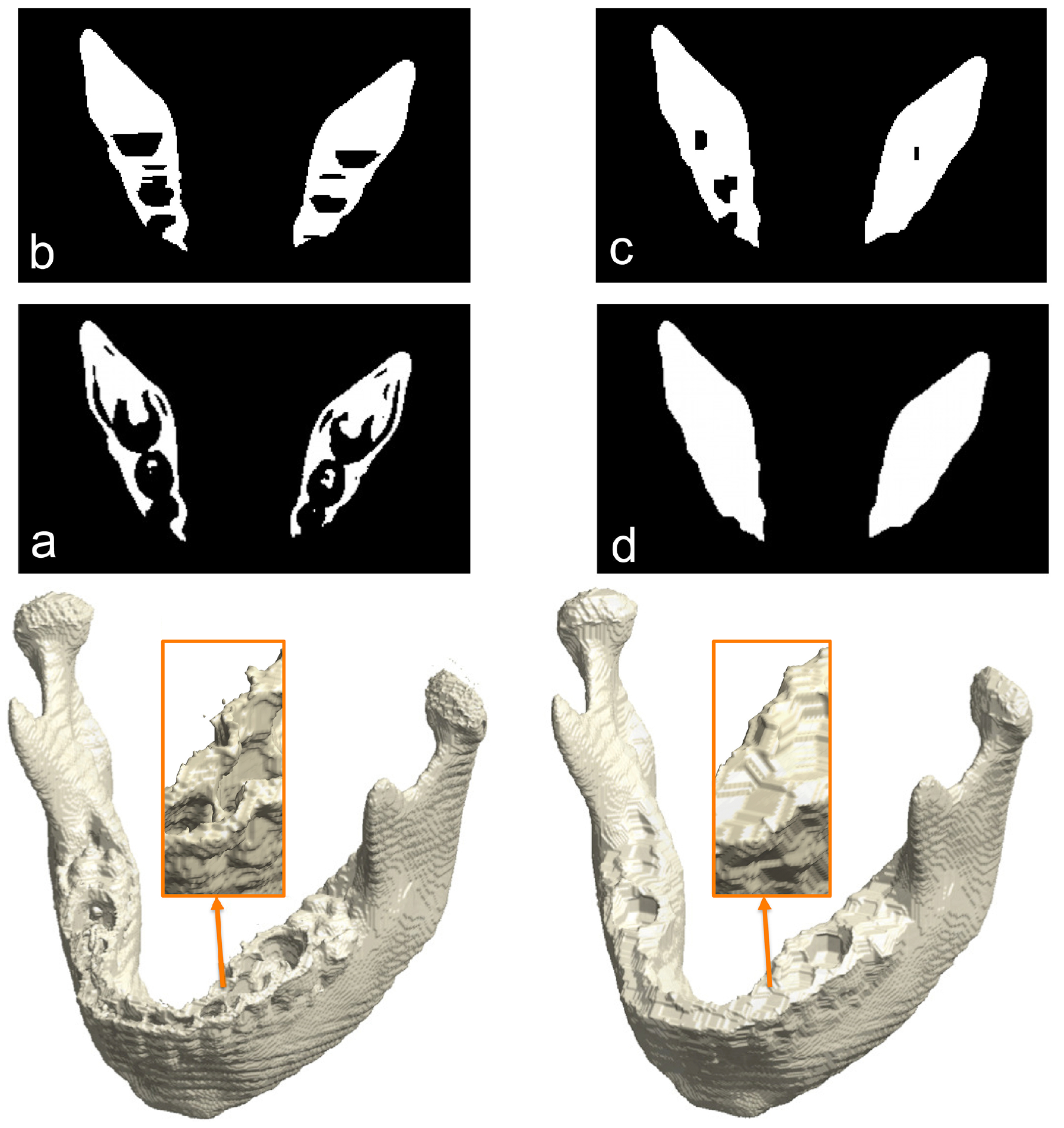}
\caption{Upper panel: Topological correction on mandible binary segmentation and surface. Disjointed tiny speckles of noisy components are removed by labeling the largest connected component, and holes and handles are removed by the 2D morphological closing operations applied sequentially to each image dimension (a $\to$ b $\to$ c $\to$ d). Lower panel left: Surface reconstruction showing holes and handles in the teeth regions. The isosurface has Euler characteristic $\chi=50$. Lower panel right: After the correction with $\chi=2$.
}
\label{fig:holepatching}
\end{center}
\end{figure}

Image acquisition and processing artifacts and partial voluming 
produce topological defects such as holes and handles in any medical image. 
In CT images of mandibles, unwanted cavities, holes  and handles in the binary segmentation mainly result from differences in CT intensity between relatively low density
mandible and teeth and more dense cortical bone and the interior trabecular bone 
 \citep{Andresen.2000.TMI, loubele.2006}. In mandibles, these topological noises can appear in thin or cancellous bone, such as in the condylar head and posterior palate \citep{stratemann.2010}. An example is shown in Figure \ref{fig:holepatching}, where the tooth cavity forms a bridge over the mandible. If we apply the isosurface extraction on the topologically defective segmentation results, the resulting surface will have many tiny handles \citep{Wood.2004,yotter.2009.MICCAI}. These handles complicate subsequent surface mesh operations such as smoothing and parameterization. It is thus necessary to correct the topology by filling the holes and removing handles. If we correct such topological defects, it is expected that the resulting isosurface is topologically equivalent to a sphere. 
 
Various topology correction techniques have been proposed in medical image processing. Rather than attempting to repair the topological defects of the already extracted surfaces \citep{Wood.2004,yotter.2009.MICCAI}, we performed  the topological simplification on the volume representation directly using morphological operations \citep{Guskov.2001,van.1992,yotter.2009.MICCAI}. The direct correction on surface meshes can possibly cause surfaces to intersect each other \citep{Wood.2004}. By checking the Euler characteristic, the holes were automatically filled up using morphological operations to make the mandible binary volume topologically equivalent to a solid sphere. All areas enclosed by the higher density bone included in the mandible definition are morphed into being in the definition of the mandible object. The hole-filled images were then converted to surface meshes via the marching cubes algorithm.

 \begin{figure}[t]
\begin{center}
\includegraphics[width=1\linewidth]{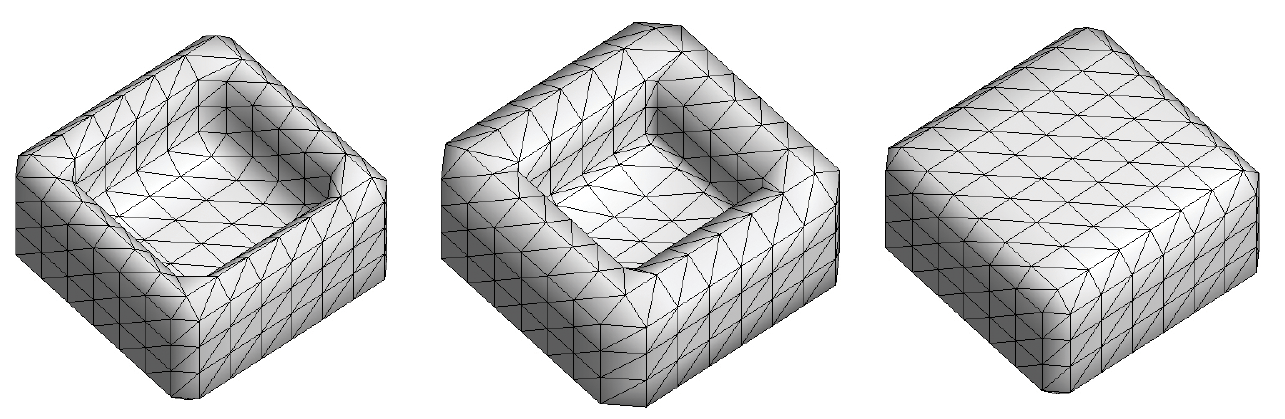}
\caption{Cavity patching by topological closing operations. Left: Surface model of the binary volume that simulates a tooth cavity. Middle: The 3D image volume based closing operation does not properly patch the cavity region. Right: The 2D image slice based closing operation patches the cavity region properly. 
}
\label{fig:holesimulation}
\end{center}
\end{figure}

In our semi-automated algorithm, we first removed the speckles of noise components by identifying the largest connected component in the binary volume. The resulting binary mandible volume was a single connected component with many small holes and handles. Then we applied the morphological closing operation in each 2D slice of CT images one by one in all three axes. Recombining the topology-corrected 2D slices resulted in topologically correct surface meshes (Figure \ref{fig:holepatching}). We used 2D topological closing operations mainly because of better performance and relatively simpler implementation than 3D topological closing operations. In 2D topological operations, we need to consider only 8 neighboring pixels compared to 26 neighboring voxels in a 3D image volume. 
There are many large concave regions left out by teeth and fillings. These regions may not be closed with a single 3D closing operation but can be easily patched up with a sequence of 2D closing operations, which put more constraints on the underlying topology. Instead of performing a single 3D closing operation that may not work, we  sequentially performed 2D closing operations in each image slice in the x- , y- and z-directions. In Figure \ref{fig:holepatching} upper panel, (a) is the binary volume before any closing operation, (b) is after the 2D closing operation in the x-direction, (c) is after the 2D closing operation in the y-direction and (d) is after the 2D closing operation in the z-direction. Each time we apply the 2D closing operation, the holes get smaller. Figure \ref{fig:holesimulation} shows a simulated cavity example that was not patched by the 3D closing operation \citep{van.1992} but was easily patched by the  sequential application of 2D closing operations. Note that any 3D object, whose every 2D cross-section is topologically equivalent to a solid disk, is topologically equivalent to a solid sphere. The problem of 3D topology correction can be thus reduced to a much simpler problem of 2D topology correction of multiple slices. Unfortunately, we cannot perform the closing operations to infinitely many possible 2D cross-sections in 3D image volumes. Therefore, we applied the 2D operations in the three axial directions. So there is a small chance the operation may not work in practice. Therefore, at the end of the processing, we performed a visual inspection of the processed volume. Further, we double checked the Euler characteristic of the resulting surface meshes. Note that for each triangle, there are three edges. For a closed surface topologically equivalent to a sphere, two adjacent triangles share the same edge. The total number of edges $E$ is thus $E = 3F/2$, where $F$ is the number of faces. If $V$ is the total number of vertices, the Euler characteristic of a sphere is given by $\chi = V- E + F = 2$. Thus, we checked if the resulting mesh satisfies the condition $V - F/2 = 2$. 77 binary mandible volumes used in the study produced the topologically correct surfaces without exception. Figure~\ref{fig:holepatching} lower panel shows an example of before and after the topology correction.

\subsection{Validation of Heat Kernel Smoothing on Mandible Surfaces}
\label{sec:validation}

The accuracy of the heat kernel construction using LB-eigenfunctions on a unit sphere using the ground truth can be found in \citet{kim.2011.PSIVT} so the results are not reproduced here. 
In this paper, we compared the performance of the proposed kernel regression against iterated kernel smoothing and diffusion smoothing using the real mandible data. The comparison results are similar for all 77 mandible surfaces used in the study, so only the results for one representative mandible surface is shown.

\begin{figure}[t]
\centering
\includegraphics[width=1\linewidth]{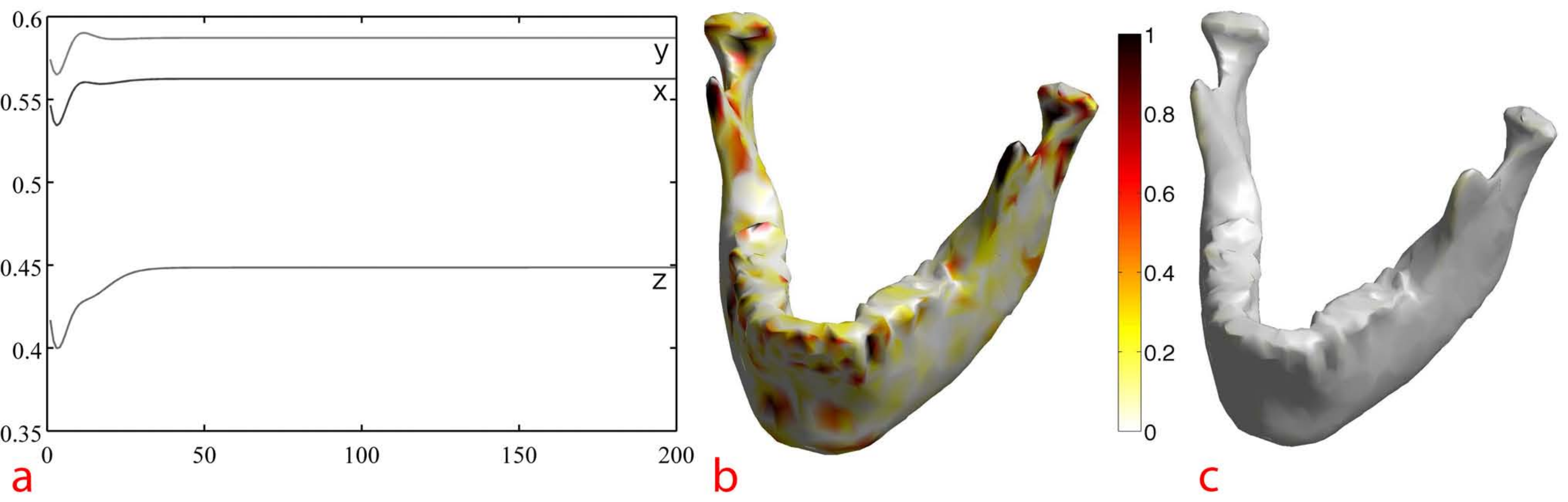} 
\caption{(a) Plot of the RMSE of the heat kernel regression against iterated kernel smoothing for coordinates $x$ (middle), $y$ (top) and $z$ (bottom) over the number of iterations up to 200. For the heat kernel regression,  $\sigma=0.5$ and $k = 132$  are used. Iterated kernel smoothing does not converge to heat diffusion. RMSE between the kernel regression and the diffusion smoothing is smaller than 0.0046 so they are not displayed in the plot. (b) The squared difference between the kernel regression and the iterated kernel smoothing. The difference is mainly localized in high curvature areas, where the Gaussian kernel used in the iterated kernel smoothing fails to approximate the heat kernel. (c) The squared difference between the kernel regression and the diffusion smoothing. There are almost no visible differences.}
\label{fig:8_relrmse}
\end{figure}

The x, y and z coordinates of a mandible surface are treated as functional measurements on the original surface and smoothed with both methods. For the comparison of performance between the smoothing methods, we calculated the root mean squared errors (RMSE) between them, where the mean of the squared errors is taken over the surface. For the heat kernel regression, we used the bandwidth $\sigma=0.5$ and eigenfunctions up to $k = 132$ degree. For iterated kernel smoothing, we varied the number of iterations $1 \leq m\leq 200$ with correspondingly smaller bandwidth $0.5/m$, which results in an effective bandwidth of $0.5$. For diffusion smoothing,  a sufficiently small  step size $\Delta \sigma = 0.0025$ was taken for 200 iterations resulting in bandwidth $\sigma =0.5$. The RMSE between the kernel regression and the iterated kernel smoothing reached up to 0.5901 ($y$-coordinate) and did not decrease even when we increased the number of iterations (Figure \ref{fig:8_relrmse}). The RMSE between the kernel regression and diffusion smoothing was smaller than 0.0046 ($y$-coordinates). Figure \ref{fig:8_relrmse} (b) shows the squared differences between the two methods. For the iterated kernel smoothing, the difference is mainly localized in high curvature areas, where the Gaussian kernel used in the iterated kernel smoothing fails to approximate the heat kernel. This comparison clearly demonstrates the limitation of iterated heat kernel smoothing, which does not converge to heat diffusion. However, the heat kernel regression and diffusion smoothing gave almost identical results and there was no discernible difference (Figure \ref{fig:8_relrmse} (c).

\begin{figure}[t]
\centering
\includegraphics[width= 1 \linewidth]{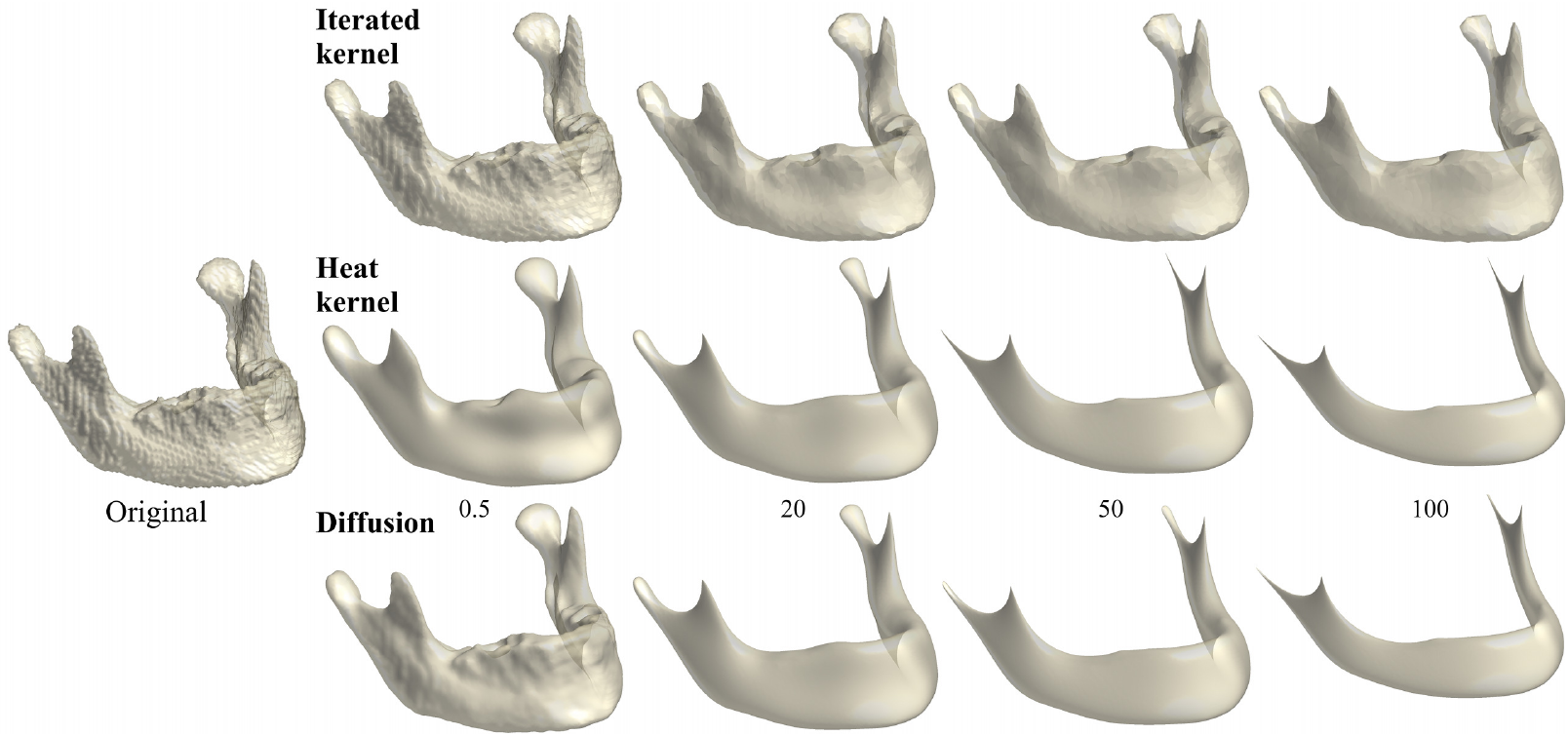} 
\caption{Smoothed mandible surfaces using three different techniques. The x, y and z surface coordinates are treated as functional measurements on the original surface and smoothed.
The proposed heat kernel smoothing is done with bandwidths, $\sigma=0.5,20,50, 100$. Iterated kernel smoothing is done by iteratively approximating the heat kernel linearly with the Gaussian kernel \citep{chung.2005.IPMI}. Diffusion smoothing  directly solves the diffusion equation using the same FEM discretization \citep{chung.2004.isbi}. Diffusion smoothing and heat kernel smoothing converge to each other as the bandwidth increases.} 
\label{fig:5_smt_surf}
\end{figure}

We also compared the performance of the three smoothing techniques at four different bandwidths $\sigma=0.5, 20, 50, 100$. For the kernel regression, $k=132$ was used. For the iterated kernel smoothing and the diffusion smoothing, a fixed step size of $\Delta \sigma = 0.025$ was used with $m=20$, 800, 2000, 4000 iterations. The diffusion smoothing and heat kernel smoothing gave visually identical results for bandwidths $\sigma = 20, 50, 100$ due to a sufficiently large number of iterations (Figure \ref{fig:5_smt_surf}). However, the iterated kernel smoothing gave a different result.
In this experiment, we replaced the original surface coordinates with smoothed ones for the final visualization. However, in the actual computation, we did not replace the original surface coordinates for the three methods. Iterated kernel smoothing compounded the discretization errors over iterations, so it did not converge to the kernel regression and diffusion smoothing. Diffusion smoothing  and heat kernel smoothing share the same FEM discretization and converge to each other as the number of iterations increases.

\subsection{Simulation Studies}
\begin{figure}[t]
\centering
\includegraphics[width=1\linewidth]{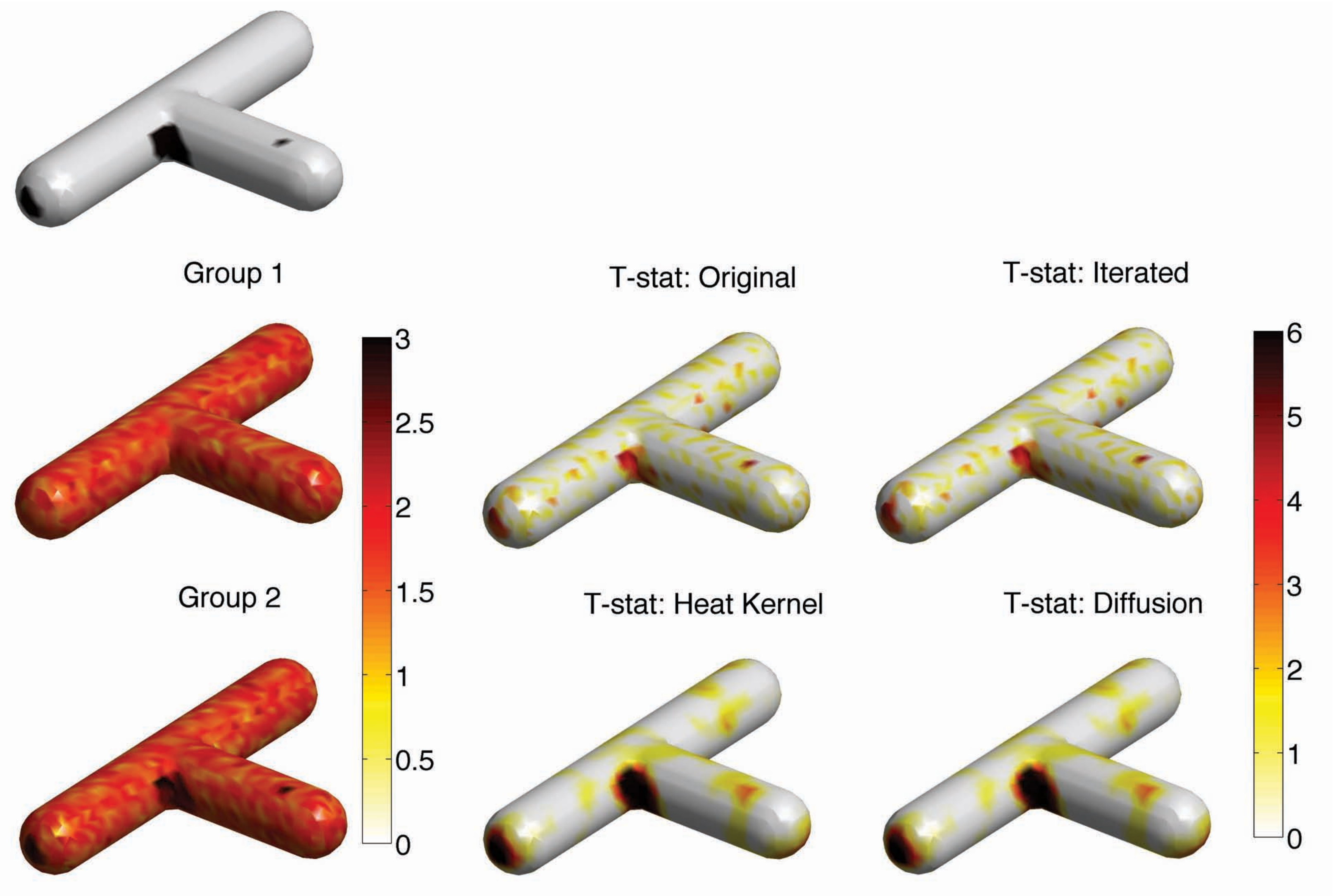} 
\caption{Simulation study I on a T-junction shaped surface where three black signal regions of different sizes are taken as the ground truth.  60 independent functional measurements on the T-junction were simulated as $| N(0, 2^2)|$  at each mesh vertex. We are only simulating positive numbers to better reflect the positive measurements used in the study. Value 1 was added to the black regions in 30  measurements, which served as group 2 while the other 30 measurements were taken as group 1.  T-statistics are shown for these simulations (original) and three techniques with bandwidth $0.5$. Heat kernel smoothing  performed the best in detecting the ground truth.}
\label{fig:8_simulation1}
\end{figure}

Since there is no known ground truth in the imaging data set we are using, it is uncertain how the proposed method will perform with real data. It is therefore necessary to perform simulation studies with ground truths. 
We performed two simulations with small and large  SNR settings on a T-junction shaped surface (Figure \ref{fig:8_simulation1}), which was chosen because it was a surface with three different curvatures: convex, concave and almost flat regions. Note that surface smoothing methods perform differently under different curvatures. Three black signal regions of different sizes were taken as the ground truth at these regions and 60 independent functional measurements on the T-junction were simulated as $|N(0,\gamma^2)|$, the absolute value of normal distribution with mean 0 and variance $\gamma^2$, at each mesh vertex. Value 1 was then added to the black regions in 30 of the measurements, which served as group 2, while the other 30 measurements were taken as group 1. Group 1 had distribution $|N(0, \gamma^2)|$ while group 2 had distribution $|N(1, \gamma^2)|$ in the signal regions. Larger variance $\gamma^2$ corresponds to smaller SNR.

\begin{figure}[t]
\centering
\includegraphics[width=1\linewidth]{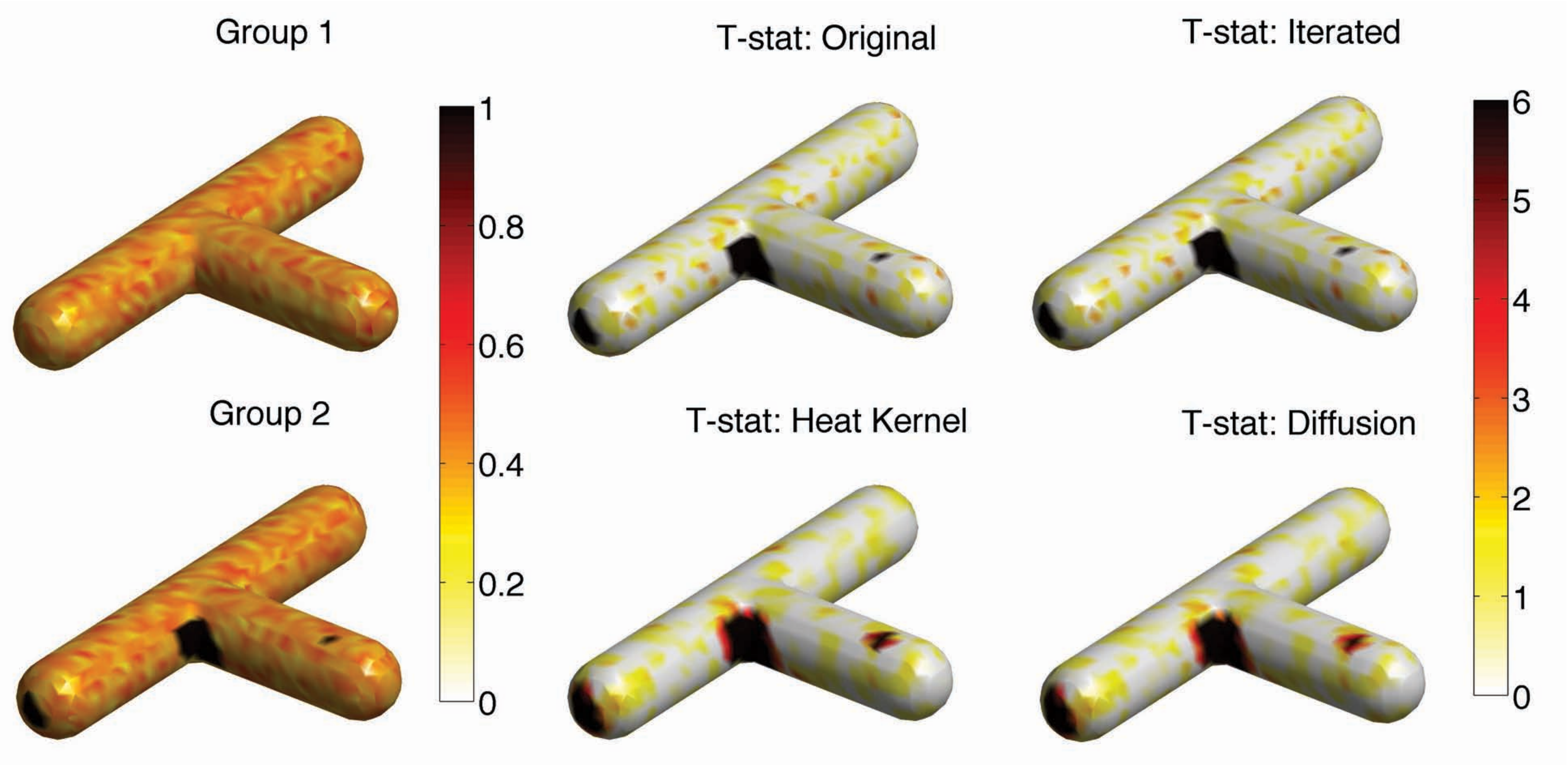} 
\caption{Simulation study II on a T-junction shaped surface with the same ground truth as simulation study I (Figure \ref{fig:8_simulation1}). 60 independent functional measurements on the T-junction were simulated as $| N(0, 0.5^2)|$  at each mesh vertex. Value 1 was added to the black regions in 30 of the measurements, which served as group 2, while the remaining 30 measurements are taken as group 1. Due to the large SNR, the group means show visible group separations. All the methods detected the signal regions; however, the heat kernel smoothing and diffusion smoothing techniques were more sensitive at the large SNR.}
\label{fig:8_simulation2}
\end{figure}

In Study I, $\gamma^2 = 2^2$ was used to simulate functional measurements with substantially smaller  SNR. Figure \ref{fig:8_simulation1} shows the simulation results. 
For iterated kernel and diffusion smoothing, we used bandwidth $\sigma = 0.5$ and 100 iterations. For smaller SNR, it is necessary to smooth with a larger bandwidth, which is determined empirically. For heat kernel smoothing, the same bandwidth and 1000 eigenfunctions were used. The same number of eigenfunctions was used throughout the study. For all three smoothing techniques, the bandwidth is the main parameter that determines performance. 
We then performed a two sample t-test with the RFT-based threshold of 4.90 to detect the group difference at $0.05$ level. 

Neither the raw data nor iterated smoothing were able to correctly identify any signal region. However, heat kernel and diffusion smoothing correctly identified 94$\%$ and 91$\%$ of the signal regions respectively. In addition, heat kernel and diffusion smoothing incorrectly identified 0.26$\%$ and 0.26$\%$ of non-signal regions as signal. There are no visually discernible differences between the two methods as shown in Figure \ref{fig:8_simulation2}. The $3\%$ difference in performance is due to the discretization error associated with taking only 100 iterations, which disappears if we take smaller time steps. Alternatively, we can use better time-discretization schemes such as Pad\'e-Chebyschev approximation \citep{patane.2015},  multi-step methods \citep{gottlieb.2001} or higher-order Runge-Kutta schemes \citep{li.2010}.

In Study II, $\gamma^2 = 0.5^2$ was used to simulate functional measurements with substantially larger SNR. Due to the large SNR, the group means showed visible group separations (Figure \ref{fig:8_simulation2}). For iterated kernel and diffusion smoothing, we used bandwidth $\sigma = 0.1$ and 100 iterations. For heat kernel smoothing, the same bandwidth and 1000 eigenfunctions were used. All the methods detected the signal regions; however, the heat kernel smoothing and diffusion smoothing techniques were more sensitive at large SNR. All the methods correctly identified the signal regions with 100$\%$ accuracy. There were no false discoveries in the raw data and iterated kernel smoothing methods. However, due to blurring effects, heat kernel and diffusion smoothing incorrectly identified 0.9$\%$ and 0.8$\%$ of non-signal regions as signal, which is negligible. For the large SNR setting, all the methods were reasonably able to detect the correct signal regions with minimal error. 

In summary, in larger SNR, all three methods performed well. However, in substantially smaller SNR, the kernel regression performed best, closely followed by diffusion smoothing. Neither the raw data nor iterated kernel smoothing performed well in the low SNR setting.

\section{Application: Mandible Growth Analysis}

As an illustration of the proposed kernel regression technique, we analyzed mandible growth on a CT imaging data set consisting of 77 human subjects between the ages of 0 and 19 years. Subjects were divided into three age categories: 0 to 6 years (group I, 26 subjects), 7 to 12 years (group II, 20 subjects), and 13 to 19 years (group III, 31 subjects). 
The main biological question of interest was whether there were localized regions of growth between these different age groups. Mandible surface meshes for all subjects were constructed through the image acquisition and processing steps described in the previous section. For surface alignment, diffeomorphic surface registration was used to match mandible surfaces across subjects \citep{miller.2009,qiu.2006.NI,qiu.2008,yang.2011}.

\subsection{Diffeomorphic Surface Registration}

\begin{figure}[t]
\centering
\includegraphics[width= 1 \linewidth]{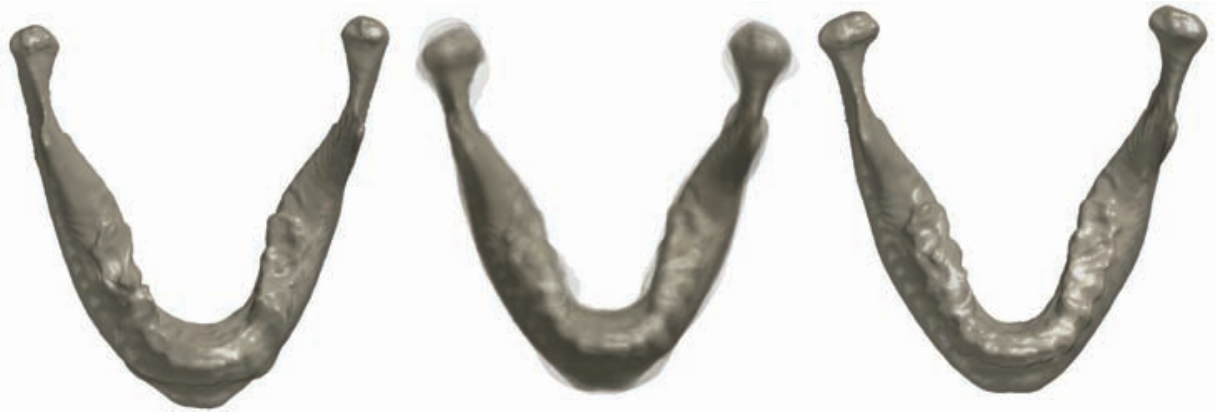} 
\caption{Left: Mandible F155-12-08, which forms an initial template $\mathcal{M}_I$. All other mandibles were affine registered to F155-12-08. Middle: The superimposition of affine registered mandibles showing local misalignment. Diffeomorphic registration was then performed to warp misaligned affine transformed mandibles. 
Right: The average of deformation with respect to F155-12-08 provides the final population average template $\mathcal{M}_F$ where statistical parametric maps were constructed.}
\label{fig:template}
\end{figure}

We chose the mandible of a 12-year-old subject identified as  F155-12-08, which served as the reference template in previous studies \citep{seo.2010.MICCAI,seo.2011.SPIE}, as initial template $\mathcal{M}_I$ and aligned the remaining 76 mandibles to the initial template affinely to remove the overall size variability. Some subjects may have larger mandibles than others, so it is necessary to remove the global size differences in localized shape modeling. From the affine transformed individual mandible surfaces $\mathcal{M}_j$, we performed an additional nonlinear surface registration to the template using the large deformation diffeomorphic metric mapping (LDDMM) framework \citep{miller.2009, qiu.2006.NI,qiu.2008,yang.2011}.

In the LDDMM framework \citep{miller.2009, qiu.2006.NI,qiu.2008,yang.2011}, 
the metric space is constructed as an orbit of surface $\mathcal{M}$ under the group of diffeomorphic transformations $\mathcal{G}$, i.e., ${\mathcal{M}}_j = \mathcal{G} \cdot \mathcal{M}$. 
The diffeomorphic transformations (one-to-one, smooth forward and inverse transformation) are introduced as transformations of the coordinates on the background space $\Omega\subset\mathbb{R}^3$.
The diffeomorphisms $\phi_t \in \mathcal{G}$ are constructed as a flow of ordinary differential equations (ODE), where $\phi_t, t \in [0,1]$ follows
\begin{align}
\dot{\phi_t} = v_t(\phi_t), \quad  \phi_0 = {\tt Id} , \quad t \in [0,1],
\label{eqn:forward-flow-equation}
\end{align}
where ${\tt Id}$ denotes the identity map and $v_t$ are the associated velocity vector fields. The vector fields $v_t$ are constrained to be sufficiently smooth, so that \eqref{eqn:forward-flow-equation} is integrable and generates diffeomorphic transformations over finite time. The smoothness is ensured by forcing $v_t$ to lie 
in a smooth vector field $V$, 
which is modeled as a reproducing kernel Hilbert space with linear operator $L$ associated with norm  
$\| u \|_V^2 = \langle Lu, u\rangle_2$ \citep{dupuis.1998}. The group of diffeomorphisms $ {\mathcal G}(V)$ is then the solutions of \eqref{eqn:forward-flow-equation} with the vector fields satisfying $\int_0^1 \| v_t \|_V dt <\infty$.

Given the template surface $\mathcal{M}$ and an individual surface $\mathcal{M}_j$, the geodesic $\phi_t, t\in [0,1]$, which lies in the manifold of diffeomorphisms and connects $\mathcal{M}$ and $\mathcal{M}_j$, is defined as 
$$
\phi_0= {\tt Id},\quad \phi_1 \cdot \mathcal{M}= \mathcal{M}_j.
$$

\begin{figure}[t]
\centering
\includegraphics[width= 1 \linewidth]{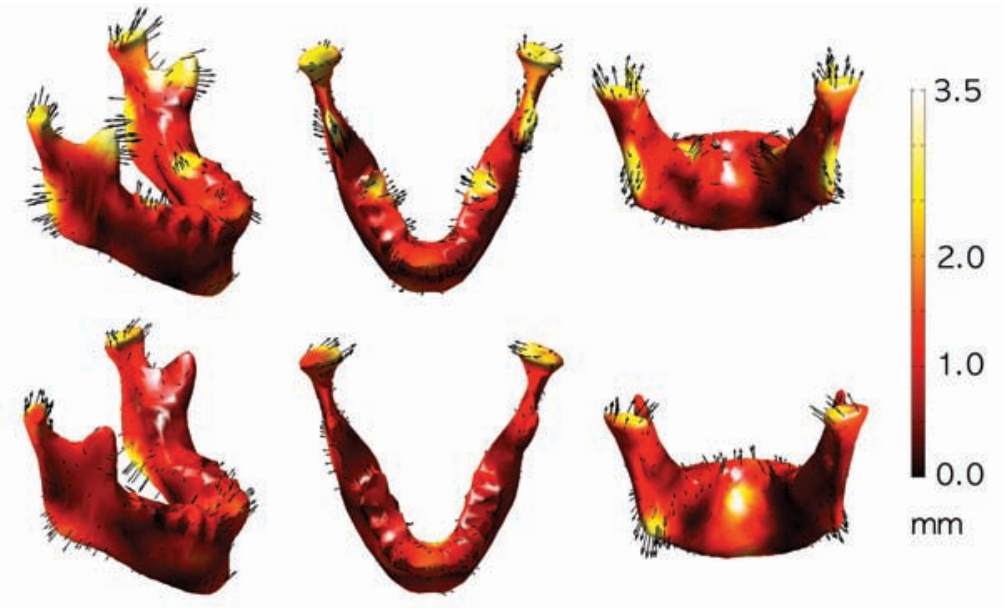} 
\caption{Mandibles were grouped into three age cohorts: group I (ages 0 to 6 years), group II (ages 7 to 12 years) and group III (ages 13 to 19 years). Each row shows the mean group differences of the displacement: group II - group I (first row) and group III - group II (second row). The arrows are the mean displacement differences and colors indicate their lengths in mm. Longer arrows imply more mean displacement.}
\label{fig:agegrouping}
\end{figure}

For our application, we employed the LDDMM approach to estimate the template among all subjects. The estimated template can be simply computed through averaging the initial velocity across all subjects \citep{zhong.2010}, which is similar to the unbiased template estimation approach in \citet{joshi.2004}. We then recomputed the displacement fields with respect to  the initial template $\mathcal{M}_I$.
We averaged the deformation fields from the initial template $\mathcal{M}_I$ to individual subjects to obtain the final template $\mathcal{M}_F$. Figure \ref{fig:template} shows the initial and final templates. Figure \ref{fig:agegrouping} shows the mean displacement differences between groups I and II (top) and II and III (bottom). Each row shows the group differences of the displacement: group II - group I (first row) and group III - group II (second row). The arrows are the growth direction with arrow length being representative of mean displacement differences and colors indicating growth length in mm.

\begin{figure}[t]
\centering
\includegraphics[width= 1 \linewidth]{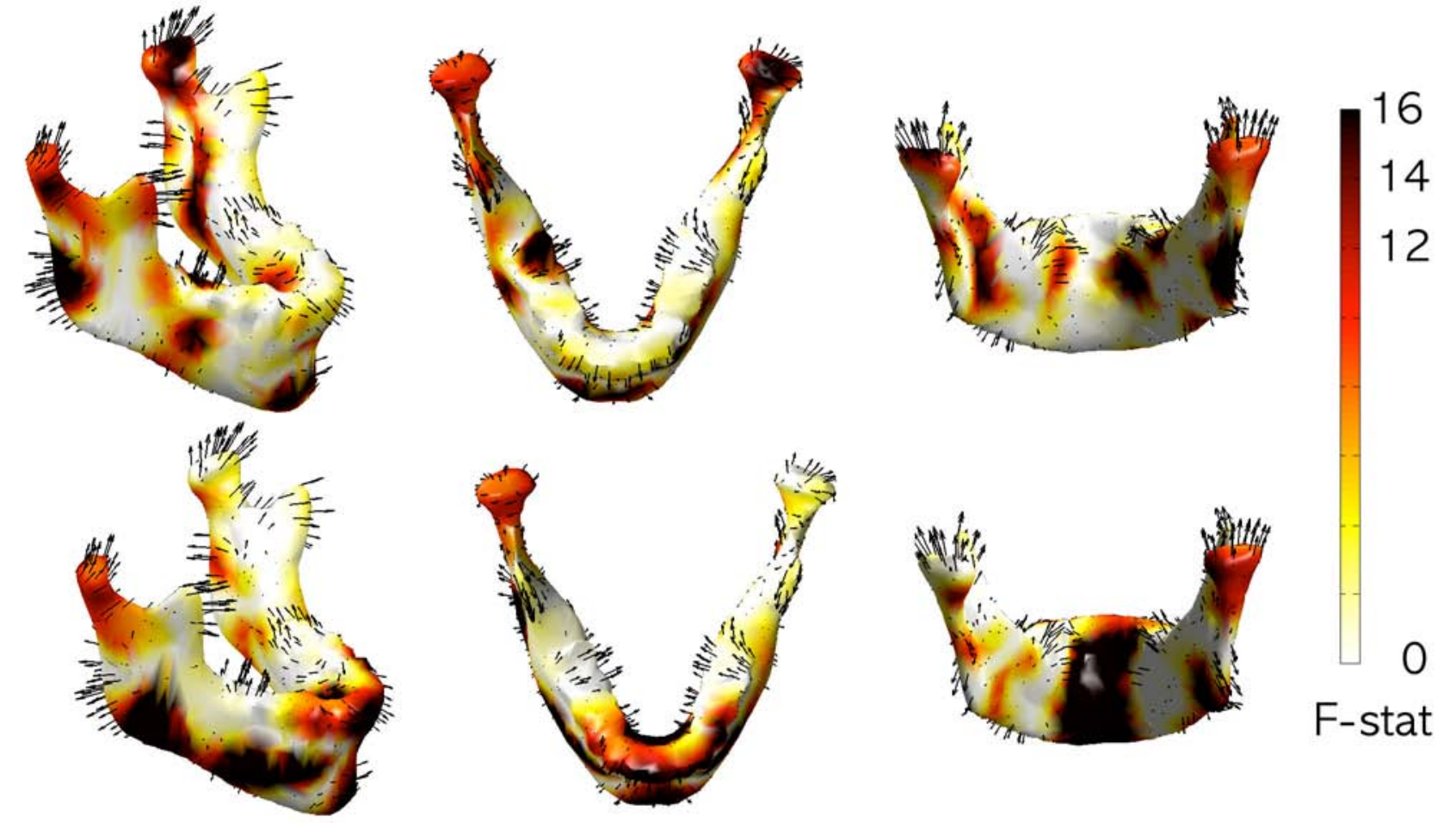} 
\caption{$F$-statistic map showing the regions of significant growth as measured by mean displacement differences between the groups displayed in Figure \ref{fig:agegrouping}. The kernel regression was used to smooth out surface measurements. The top row shows significant growth between groups I and II ; and bottom row between groups II and III. The thresholds 10.52 and 10.67 are considered significant at 0.01 level (corrected) for the top and bottom rows.}
\label{fig:Fmap}
\end{figure}

\begin{figure}[t]
\centering
\includegraphics[width= 1 \linewidth]{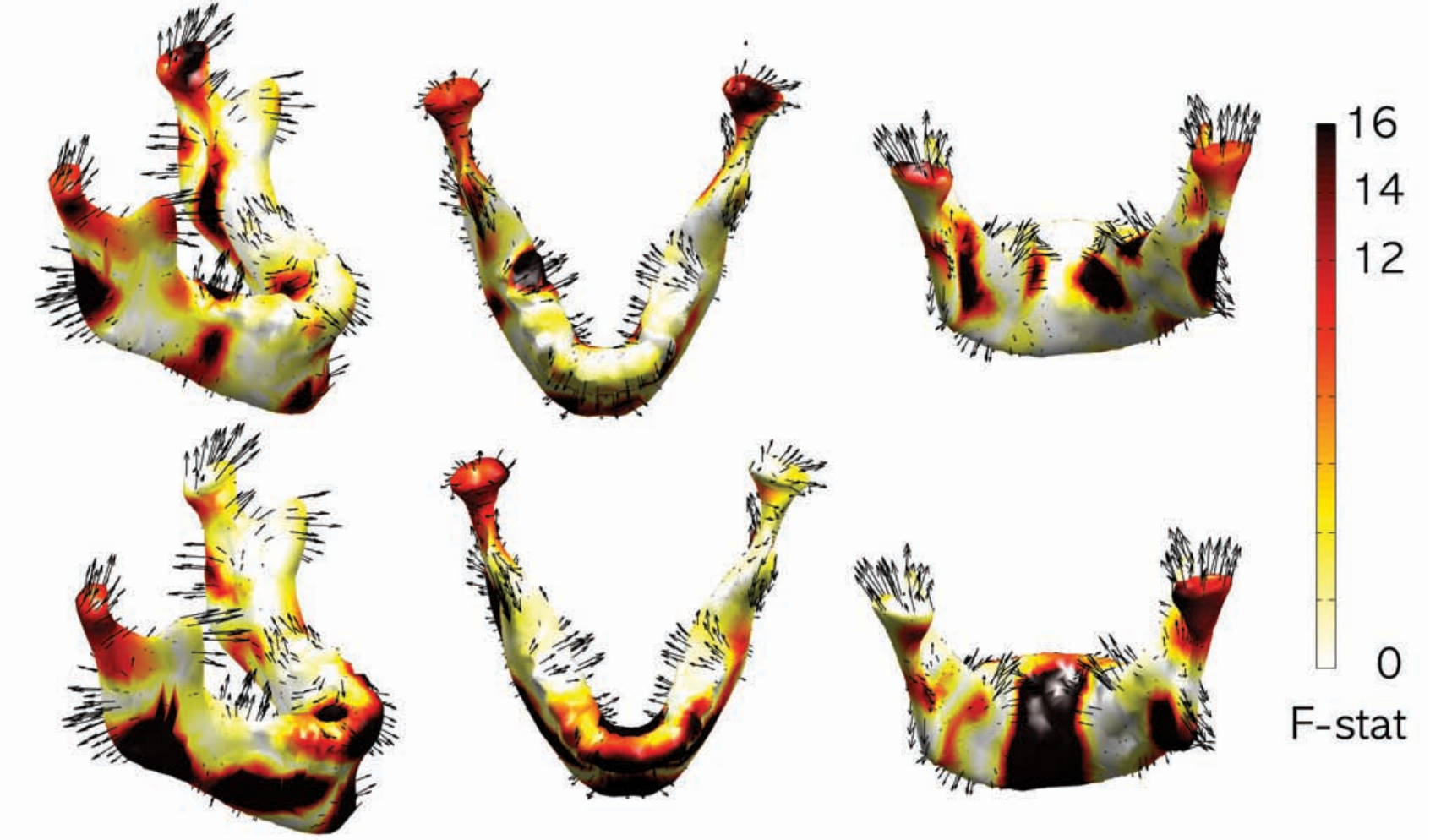} 
\caption{$F$-statistic map showing the regions of significant growth as measured by mean displacement differences between the groups displayed in Figure \ref{fig:agegrouping}. The iterated kernel smoothing with parameters $\sigma=20$ and $m=200$ were used. The top row shows significant growth between groups I and II ; and bottom row between groups II and III. The thresholds 10.52 and 10.67 are considered significant at 0.01 level (corrected) for the top and bottom rows.}
\label{fig:Fmapiternated}
\end{figure}

\subsection{Statistical Analysis}
We are interested in determining the significance of the mean displacement differences in Figure \ref{fig:agegrouping}. Since  the length measurement provides a much easier biological interpretation, we used the length of the displacement vector as a response variable. 
The RFT assumes the measurements to be a smooth Gaussian random field. Heat kernel smoothing on the length measurement will make the measurement smoother, more Gaussian and increase the SNR \citep{chung.2005.IPMI}. Heat kernel smoothing is applied with bandwidth $\sigma =20$ using $1000$ eigenfunctions on the final template $\mathcal{M}_F$. The number of eigenfunctions used is more than sufficient to guarantee a relative error less than $0.3\%$. The heat kernel smoothing of the displacement length is taken as the response variable. We constructed the $F$ random field testing the length difference between the age groups I and II, and II and III showing the regions of accelerated growth (Figure \ref{fig:Fmap}). 

The comparison of  groups I and II is based on an $F$-field with 1 and  44 degrees of freedom. The comparison of groups II and III  is based on an $F$-field with 1 and  49 degrees of freedom. The multiple comparison corrected $F$-statistics thresholds corresponding to $\alpha = 0.05$ and $0.01$ levels are respectively 8.00 and 10.52 (group II-I) and 8.00 and 10.67 (group III- II). In the $F$-statistic map shown in Figure \ref{fig:Fmap}, black and red regions are considered as exhibiting growth spurts at 0.01 and 0.05 levels respectively.  Our findings are consistent with previous findings of simultaneous forward and downward growth \citep{scott.1976,smartt.2005, walker.1972,lewis.1982,seo.2011.SPIE} and bilateral growth \citep{enlow.1996}.

We also performed the same statistical analysis to the iterated kernel smoothing and diffusion smoothing results. For the diffusion smoothing, 200 step sizes are used with the fixed time step $0.01$, which results in the overall bandwidth $\sigma=20$.  For the iterated kernel smoothing, bandwidth $\sigma=20$ is split into $m=200$ small bandwidths. The diffusion smoothing results are similar to Figure \ref{fig:Fmap} so the only iterated kernel smoothing result is shown in Figure \ref{fig:Fmapiternated}. Since this is a high SNR setting, all three methods are expected to perform well and similarly. In the heat kernel regression, 25$\%$ of mesh vertices show $F$-statistic value above 8.00  for the comparison of groups I and II (0.05 level). For the iterated kernel smoothing and diffusion smoothing, 24$\%$ and 24$\%$ of vertices are above 8.00. For the comparison of groups II and III, the numbers are 38$\%$, 36$\%$ and 36$\%$ respectively. The differences are not significant.

\section{Conclusions}

This study presents a novel heat kernel regression framework, where 
functional measurements are expanded analytically using the weighted Laplace-Beltrami eigenfunctions. 
The weighted eigenfunction expansion is related to isotropic heat diffusion and the diffusion wavelet transform. The method was validated and compared against exiting surface-based smoothing methods. 
Although the proposed kernel regression and diffusion smoothing share 
identical FEM discretization, the kernel regression is a parametric model, whereas diffusion smoothing is not. The flexibility of the parametric model  enabled us to establish the mathematical equivalence of kernel regression, diffusion smoothing and diffusion wavelets.

The method was subsequently applied to characterize mandible growth.
Based on the significant directions of growth identified in Figure \ref{fig:agegrouping} and \ref{fig:Fmap}, we 
quantified the regions, direction and extent of  growth during 
the first two decades of life that contribute to the overall downward and forward growth of the mandible as described in the literature. To quantify mandibular growth using smaller  age cohorts, we are currently securing additional CT images.

\section*{Acknowledgment}
This work was supported by NIH Research Grant R01 DC6282 (MRI and CT Studies of the Developing Vocal Tract), from the National Institute of Deafness and other Communicative Disorders (NIDCD);  core grant P-30 HD03352 to the Waisman Center from the National Institute of Child Health and Human Development (NICHHD); Clinical and Translational Science Award (CTSA) program, through the NIH National Center for Advancing Translational Sciences (NCATS) UL1TR000427, and Vilas Associate Award from University of Wisconsin-Madison. We thank Lindell R. Gentry, Mike S. Schimek, Brian J. Whyms, Reid B. Durtschi and Dongjun Chung at the University of Wisconsin-Madison for assistance with image acquisition and image segmentation. We also thank the anonymous reviewers and Yuan Wang and Jacqueline Houtman for comments on earlier versions of this paper.

\bibliographystyle{model2-names}
\bibliography{reference.2015.01.30} 
\end{document}